\documentclass[final]{cvpr}

\usepackage{times}
\usepackage{epsfig}
\usepackage{graphicx}
\usepackage{amsmath}
\usepackage{amssymb}
\usepackage{stmaryrd}
\usepackage{caption}
\usepackage{subcaption}

\usepackage{amsmath}
\usepackage{amssymb}
\usepackage{amsthm}
\usepackage{mathtools}

\DeclareMathOperator{\softmax}{softmax}

\usepackage[pagebackref=true,breaklinks=true,colorlinks,bookmarks=false]{hyperref}

\def\cvprPaperID{5598} % *** Enter the CVPR Paper ID here
\def\confYear{CVPR 2021}

\begin{document}

%%%%%%%%% TITLE
%\title{\CVPR21, Attention-Fewshot Paper}
\title{SAFCAR: Structured Attention Fusion for Compositional Action Recognition}

\author{Tae Soo Kim   \qquad \qquad  Gregory D. Hager \\
Johns Hopkins University\\
3400 N. Charles St, Baltimore, MD\\
{\tt\small \{tkim60,hager\}@jhu.edu}
% For a paper whose authors are all at the same institution,
% omit the following lines up until the closing ``}''.
% Additional authors and addresses can be added with ``\and'',
% just like the second author.
% To save space, use either the email address or home page, not both
% \and
% Gregory D. Hager\\
% Johns Hopkins University\\
% First line of institution2 address\\
%{\tt\small secondauthor@i2.org}
}

\maketitle

%%%%%%%%% ABSTRACT
\begin{abstract}
   % When an action is `compositional' such that it is defined by a verb-noun pair (ie. Placing a pen), a combinatorial number of unique instantiations of the same action category exists when composed with different objects. This makes enumeration of all possible combinations for training impractical. Inevitably, a trained model faces unseen verb-noun compositions at test time in realistic applications. Thus, a successful compositional approach must be able to capture the essential spatial-temporal structure of an action that generalizes across to unseen compositions. In this paper, we present a generic framework for recognizing compositional actions using an approach with two pathways: one for learning the spatial-temporal structure of an action using object detections and one for extracting useful visual cues from video. We introduce a Structured Attention Fusion (SAF) mechanism to effectively fuse the representation of the two pathways using a cross-pathway attention operator. We evaluate our method on the compositional as well as the challenging few-shot compositional split (the Something-Else tasks) of the Something-Something-V2 dataset  where we establish a new state of the art on both tasks. Finally, we show that our model generalizes to a different domain of actions found in the Charades dataset where again establish a new state of the art of the few-shot task.
%In compositional action recognition, a combination of primitive components defines an action (ie. Place a pen). 
We present a general framework for compositional action recognition -- i.e. action recognition where the labels are composed out of simpler components such as subjects, atomic-actions and objects.
The main challenge in compositional action recognition is that there is a combinatorially large set of possible actions that can be composed using basic components. However, compositionality also provides a structure that can be exploited. To do so, we develop and test a novel Structured Attention Fusion (SAF) self-attention mechanism to combine information from object detections, which capture the time-series structure of an action, with visual cues that capture contextual information.  
We show that our approach recognizes novel verb-noun compositions more effectively than current state of the art systems, and it generalizes to unseen action categories quite efficiently from only a few labeled examples. We validate our approach on the challenging Something-Else tasks from the Something-Something-V2 dataset. We further show that our framework is flexible and can generalize to a new domain by showing competitive results on the Charades-Fewshot dataset.

\end{abstract}

%%%%%%%%% BODY TEXT
\section{Introduction}

Descriptions of human action can be naturally composed from basic components such as subjects (nouns), atomic-actions (verbs) and objects. Automated recognition of human actions in video thus faces the fundamental challenge that the set of labels is combinatorially large. As a result, this makes enumeration of all possible descriptions to train end-to-end 3D convolutional methods \cite{tsm,Carreira2017QuoVA,feichtenhofer2018slowfast,lfb2019,hussein2018timeception} impractical in this domain. Instead, a model must have a means to represent and ``assemble'' the spatial-temporal structure of an action so that it can generalize to novel instances of actions composed with different and possibly previously unseen components. 
\begin{figure}[t]
\centering
\includegraphics[width=1.0\linewidth]{ 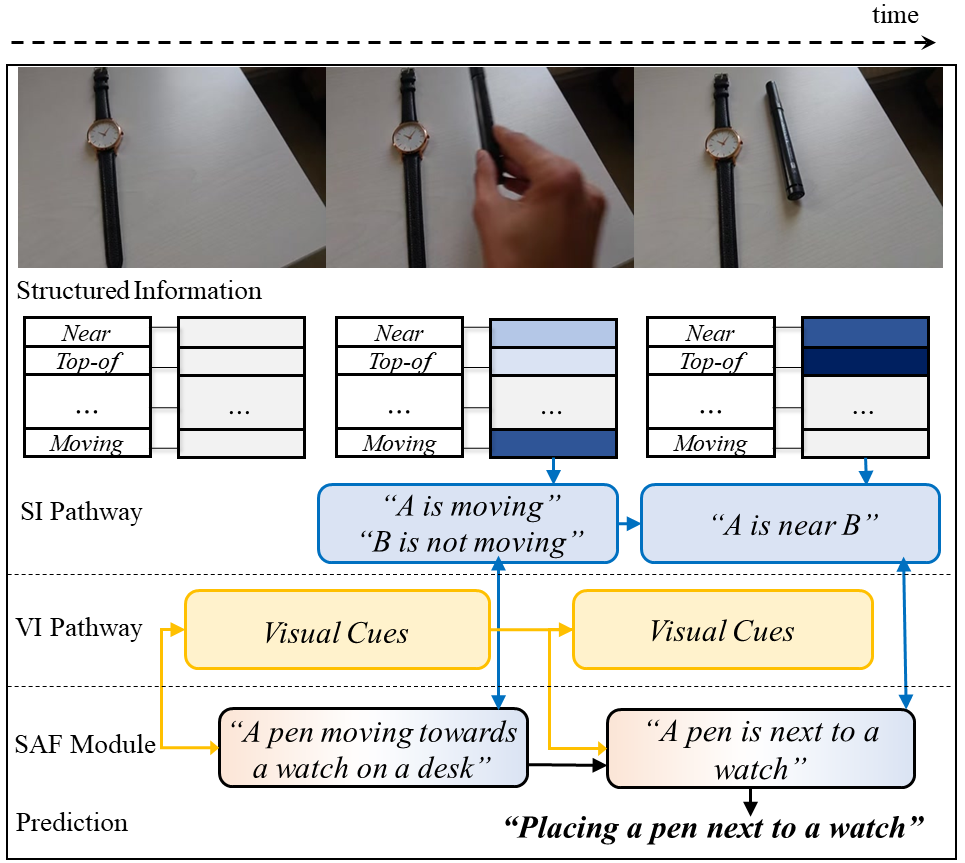}
  \caption{Our framework uses two pathways (Structured and Visual Information pathways) to learn structured components of an action. Structured Information (SI) pathway captures structure of an action using object detection. We show that the Structured Attention Fusion mechanism allows the rich information embedded in a video to refine the structured representation learned in the SI pathway.}
\label{fig:thumbnail}
\end{figure}

%% Many methods haven proposed in the literaure, 1,2,3. However these methods {limitations}...  End with what is missing

Many methods have been proposed to model the spatial-temporal structure of actions using objects. In such methods, the structure of an action is approximated using variations of specified space-time graphs \cite{Materzynska_2020_CVPR,Jain2016StructuralRNNDL}, learned using a graph convolutional network \cite{Wang_videogcnECCV2018} or manually specified by annotating the relations between objects \cite{Ji_2020_CVPR}. However, such methods do not exhibit clear and obvious improvements unless ensembled with end-to-end RGB baselines \cite{Materzynska_2020_CVPR} at test time or ground truth knowledge about actor-object relations is provided by an oracle \cite{Ji_2020_CVPR}. Findings suggest that the structured information learned using objects and the visual information are not fused effectively.

In this paper, we build on the idea that action structure learning using object detections can be synergistically coupled with the rich visual cues available in video. This leads to a generic framework for effectively recognizing compositional actions using two pathways. First, the Structured Information (SI) pathway explicitly reasons about the  spatial-temporal structure of an action from object detections. Using a temporal model such as a temporal convolutional network \cite{tcn,restcn}, the SI pathway learns to encode spatial relations between actors as well as consistent temporal patterns of those relations. Second, the Visual Information (VI) pathway extracts a time-series of visual features from the images comprising the video. 

A common way to fuse object-level and visual information is via a naive concatenation or ensembling \cite{Materzynska_2020_CVPR}. Instead, we
introduce the Structured Attention Fusion (SAF) mechanism which allows both pathways to attend to each other and fuse the structural and visual information from the two pathways. We show that we can learn structures of an action more effectively by allowing both pathways to attend to each other to refine the fused representation. Figure \ref{fig:thumbnail} illustrates the interplay of the two pathways through the SAF module. For example, when only using 2D object detections as inputs to the SI-pathway, fine-grained spatial structure between actors such as whether an object is next to, inside or above another object may be ambiguous. By allowing the structured representation to attend to the visual cues from the VI-pathway, we can eliminate such ambiguities as illustrated in Figure \ref{fig:thumbnail}. 

In the remainder of this paper, we show that our approach significantly improves on current methods for compositional action recognition. We demonstrate that our approach can generalize over object appearance while preserving the core spatial-temporal structure of an action. With this ability, our approach can recognize unseen verb-noun compositions at test time better than other approaches and establishes a new state of the art on the Something-Else task \cite{Materzynska_2020_CVPR} from the Something-Something-V2  dataset \cite{something}. In addition to novel verb-noun compositions, we show that the learned representation generalizes to unseen action categories (verbs) efficiently using only a few labeled examples. We verify this by evaluating on the very challenging fewshot compositional split proposed by \cite{Materzynska_2020_CVPR} where we again establish a new state of the art. Finally, we show that our model generalizes to a different domain of actions found in the Charades dataset \cite{sigurdsson2016hollywood} where our approach performs competitively on the Charades-Fewshot benchmark \cite{Ji_2020_CVPR}.
In summary, the main contributions of the paper are:
\begin{enumerate}
    \item A general framework for compositional action recognition using multiple pathways with the SAF module.
    \item A new state-of-the-art for recognizing compositional actions with unseen components.
    \item A representation of an action that requires fewer training examples to adapt to unseen actions, validated across multiple domains including the Something-Else and Charades-Fewshot tasks.
\end{enumerate}

\section{Related Work}
\textbf{Action Classification in Videos:}
With the introduction of large scale video datasets for action classification \cite{caba2015activitynet,ava,kinetics}, many deep neural architectures have been proposed to extract powerful representations from videos \cite{bourdev,Carreira2017QuoVA,feichtenhofer2018slowfast,vgg_twostream,hussein2018timeception,tsm,c3d,lfb2019}. Most approaches for video classification pretrain such models on large-scale datasets and later finetune to target domains. However, the findings in \cite{trn} suggest that such pretrained models focus more on appearance rather than the temporal structure of actions. To evaluate whether a video model can generalize over appearance, researchers have proposed more fine-grained benchmarks and structured tasks \cite{something,Ji_2020_CVPR,Materzynska_2020_CVPR,sigurdsson2016hollywood} including compositional action recognition. In this work, we focus on compositional and few-shot compositional action recognition tasks where a model must be able to generalize to actions defined by novel verb-noun compositions.

\textbf{Few-shot Learning}:
The few-shot setting challenges a model's ability to generalize to unseen examples; we additionally validate our approach in the few-shot setting. Recent work \cite{chen2019closerfewshot} showed that a complex tasks/episodes-based meta-learning approach \cite{Bishay2019TARNTA,NIPS2016_6385,NIPS2017_6996,compare,46678,Dwivedi2019ProtoGANTF} is not necessary when a good initialization of a model is combined with a distance based classifier \cite{finn,oneshot,mishra,chen2019closerfewshot}. We demonstrate with our few-shot experiments that our approach leads to a good initialization such that it only requires few examples to transfer to novel action categories. We show in our experiments that our approach has learned structural components of an action which can adapt quickly to novel categories using only a handful of labeled instances.

\textbf{Use of Attention: }
Attention mechanisms have been widely adopted in many static image-based  \cite{hu2018genet,squeeze,Woo_2018_BMVC, Woo_2018_ECCV,Gao_2019_CVPR} and video-based \cite{Du_2018_ECCV,squeeze,A2_nips,staa,perezrua2020knowing,wang2020eca,Bishay2019TARNTA} classification problems. The role of the attention mechanism can be interpreted as a way to implicitly focus feature extraction on important aspects of an action. In existing methods, this process happens within a single pathway using only RGB input. We use the attention mechanism to fuse the information \textit{between} pathways operating on different input modalities.

In particular, the recently introduced self-attention \cite{Vaswani_attention} mechanism captures long-range dependencies within data points; it has quickly become the core building block of state of the art approaches for multiple natural language modeling tasks  \cite{brown2020language,devlin-etal-2019-bert,roberta,NIPS2019_8812}. In the vision domain, recent papers have investigated how self-attention can generalize to the image domain \cite{AAconv,standalone,Cordonnier2020On} as well as to video-based applications \cite{NL_2018_CVPR}. The Non-local Neural Network \cite{NL_2018_CVPR} captures long-range dependencies within a video by use of a non-local operator which is a generalization of the self-attention unit \cite{Vaswani_attention}. Our key observation is that the attention operation leads to a fusion of information whether it is within the model's spatial-temporal feature maps \cite{NL_2018_CVPR} or between long-term and short-term features \cite{lfb2019} in a feature-bank framework. We investigate its use in fusing structural and visual information of an action in a multimodal framework.

%specifically for static applications such as image classification and object detection, there exists recent work that investigates how the convolution operator in modern convolutional neural networks can be augmented with the self-attention operator \cite{AAconv} or how convolution layers can be replaced entirely by the self-attention layers \cite{standalone,Cordonnier2020On}. For video based action classification, Non-local Neural Networks \cite{NL_2018_CVPR} haven shown to effectively capture long-range dependencies in videos by use of a non-local operator which is a generalization of the self-attention unit from \cite{Vaswani_attention}. Our approach is closely related to recent work and we investigate whether the non-local (self-attention) operators can be used to implicitly learn more structured representation of actions by fusing the attentions maps learned from multiple inputs such as videos, object detections and scene graph predictions.

\textbf{Structured Representations of Videos using Objects:} A growing line of work uses structured information extracted from videos, such as object detections and scene graphs, to improve fine-grained analysis of actions. The most common form of structure in action classification is obtained through detectable objects and actors in the scene \cite{Baradel_2018_ECCV,Jain2016StructuralRNNDL,girdhar2019video, lfb2019,Wang_videogcnECCV2018,attendinteract,Sun2018ActorCentricRN,Ji_2020_CVPR,Materzynska_2020_CVPR}. Instead of learning features using only video, these approaches often combine features extracted from regions of interest using detections provided by state of the art object detectors \cite{maskrcnn}. The object centric representations can then be used to learn pairwise relations between objects \cite{Materzynska_2020_CVPR,attendinteract,Baradel_2018_ECCV}, between objects and global context \cite{lfb2019,Sun2018ActorCentricRN,girdhar2019video,Materzynska_2020_CVPR} and within a specified graph structure \cite{Wang_videogcnECCV2018,Ji_2020_CVPR} to improve action classification. The object-centric information can then be used to predict action labels directly \cite{Baradel_2018_ECCV,attendinteract,Sun2018ActorCentricRN,girdhar2019video,Materzynska_2020_CVPR} via end-to-end training. However, such methods do not exhibit clear and  obvious improvements unless ensembled with end-to-end RGB  baselines at test time \cite{Materzynska_2020_CVPR} or  ground truth knowledge about actor-object relations is provided \cite{Ji_2020_CVPR}. Findings suggest that the structured information learned using objects and the visual information are not fused effectively. We address this issue by using an attention based SAF module to fuse structural and visual information. This is similar to the recent Feature-Bank-Operator \cite{lfb2019} where long-term ROI-aligned object-centric features \cite{faster_rcnn} are merged with short-term visual features from 3D CNNs. However, the information being fused with attention is fundamentally different between the approaches. In \cite{lfb2019}, visual features extracted around humans or ROI-aligned object features are used directly without extracting high-level structure from them. We extract high-level action structures from objects using the SI pathway which get fused with the visual information.

\textbf{Architecture:} The well known two-stream architecture \cite{vgg_twostream} has been modified and adopted in state-of-the-art video recognition models such as \cite{Carreira2017QuoVA,feichtenhofer2018slowfast}. Two stream architectures have shown to be effective when learning with optical-flow + RGB inputs \cite{vgg_twostream,Carreira2017QuoVA} and when merging RGB videos with different temporal speeds \cite{feichtenhofer2018slowfast}. We adopt a two stream architecture that  leverages object detection in one stream and raw pixels in the other pathway. We empirically show that fusing information between the two pathways through the SAF module greatly improves performance in compositional action recognition settings.

\begin{figure*}
     \centering
     \begin{subfigure}[b]{0.62\linewidth}
         \centering
         \includegraphics[width=\linewidth]{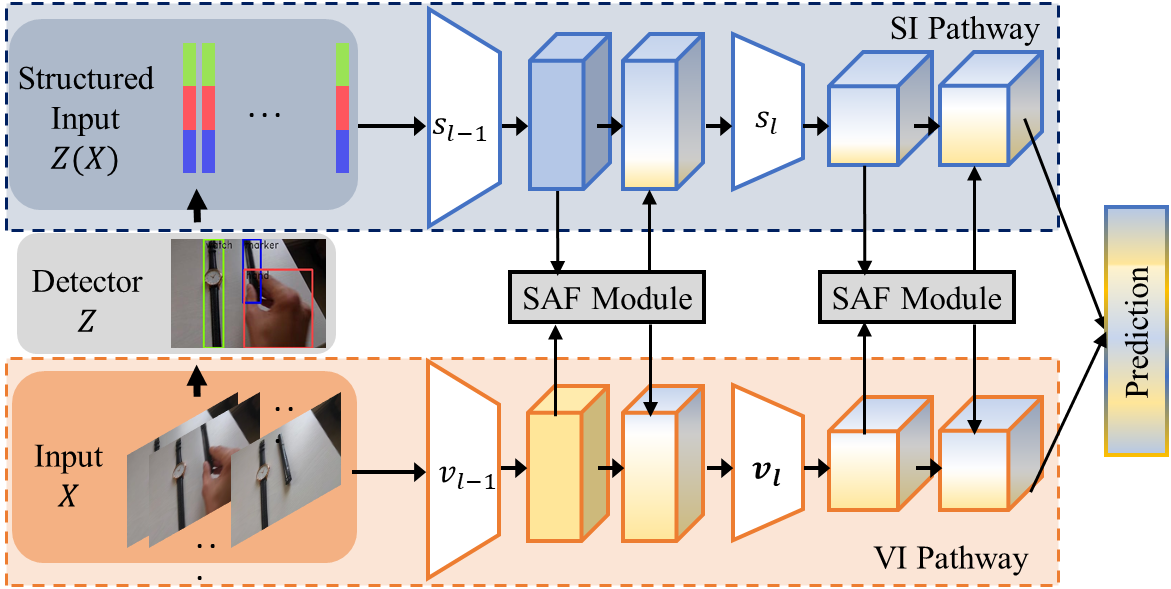}
         \caption{An example instantiation of our framework.}
         \label{fig:architecture}
     \end{subfigure}
     \hfill
     \begin{subfigure}[b]{0.35\linewidth}
         \centering
         \includegraphics[width=\linewidth]{ 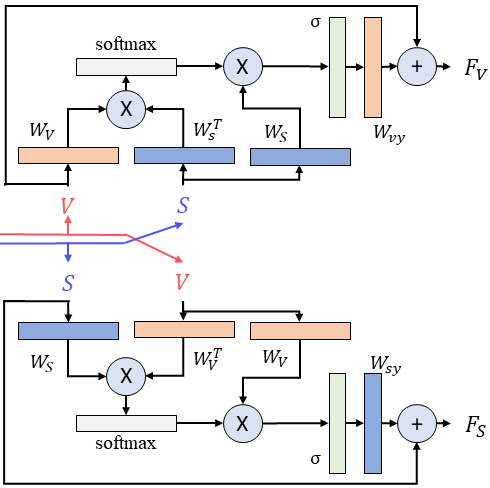}
         \caption{Attention based SAF module. }
         \label{fig:saf}
     \end{subfigure}
     
    \caption{Figure \ref{fig:architecture} is an example instantiation of our framework which accepts objects detections and a video as inputs to their corresponding pathways. We plot the Structured Attention Fusion module in Figure \ref{fig:saf}. The inputs to the module are intermediate representation $S$ (Structured) and $V$ (Visual) from the respective pathways.}
    \label{fig:overall}
\end{figure*}
\section{SAFCAR: Structured Attention Fusion for Compositional Action Recognition}
%When actions are compositional and fine-grained, a successful vision based action classifier must be able to recognize relevant spatial-temporal relations between structured components of the scene and generalize over potential nuisance factors such as changes in object appearance, identity, layout and context. 
%A successful model for compositional action recognition  must be able to recognize  spatial-temporal structure of an action. 
A successful model for compositional action recognition  must abstract out structure of an action from appearance of its components. To that end, we use a dedicated Structured Information (SI) (Sec. \ref{sec:si}) pathway to explicitly learn the structure of an action using objects and a Visual Information (VI) (Sec. \ref{sec:vi}) pathway to learn useful complementary visual concepts from a video. Structured Attention Fusion (SAF)  (Sec. \ref{sec:SIF}) mechanism fuses the representations learned from the two pathways by allowing them to attend to each other. Figure \ref{fig:architecture} illustrates an example instantiation of the framework where object detections are used in the SI pathway.

\subsection{Visual Information (VI) Pathway}
\label{sec:vi}
Let $X \in \mathbb{R}^{T \times H \times W \times C}$ be a video with $C$ channels with $T$ frames with spatial dimension $H$ and $W$. The VI pathway can be any feed-forward neural architecture $V$ that has the following form:
\begin{equation}
V(X) = v_L(v_{L-1}(\dots v_2(v_1(X)))) 
\end{equation}
where the $l$-th intermediate representation is computed by sub-modules $v_{1:l}$ of the network and $V(X) \in \mathbb{R}^{T_V^L \times H_V^L \times W_V^L \times C_V^L }$. There are many well-established convolutional neural architectures that satisfy the above conditions \cite{Carreira2017QuoVA,tsm,feichtenhofer2018slowfast,vgg_twostream} and our general framework supports the use of any such models. For classification problems, a prediction $\hat{y}$ is made using a learned classification head $v_{cls}:  \mathbb{R}^{C_V^L} \shortrightarrow \mathbb{R}^{N}$:
\begin{equation}
\hat{y} = \softmax(v_{cls}(\sigma(V(X))))
\end{equation}
where $N$ is the number of classes and $\sigma$ is an aggregation function such as a global pooling operator over both spatial and temporal dimensions of $V(X)$. When trained sufficiently, we know that $V(X)$ contains high-level visual features such as movement patterns, context and other visual cues useful for the action classification task.

\subsection{Structured Information (SI) Pathway}
\label{sec:si}
The goal of the SI pathway is to extract structured information regarding the spatial-temporal components of an action. Let $Z(X) \in \mathbb{R}^{D  \times T}$ be some structured representation of the video $X$. As illustrated in Figure \ref{fig:architecture}, a concrete example of $Z(X)$ may be a sequence of per-frame object detections provided by some object detector $Z$. Given a time-series of frame-level structural information, let $S$ be a feed-forward neural architecture such that :
\begin{equation}
S(X) = s_L(s_{L-1}(\dots s_2(s_1(Z(X))))) 
\end{equation}
where $s_l$ is a submodule of $S$, $1 \leq l \leq L$  and $S(X) \in \mathbb{R}^{T_S^L \times  C_S^L}$. Our framework does not require the depth of the $V$ and $S$ to be the same; it only requires that at least one component of $s_l$ contains a neural operator with learnable parameters that performs temporal feature extraction. For example, each $s_l$ can be implemented as a temporal convolution layer with 1D convolutions followed by a non-linear operation \cite{Lea2017TemporalCN}, a recurrent layer such as \cite{lstm,gru}, a self-attention based transformer encoder \cite{Vaswani_attention} or any mixture of such components. %When trained end-to-end using object detections,
We show in our experiments that the SI-pathway alone can outperform complex video models. We describe our approach for fusing the representation learned using both pathways.
%while using an object detector to construct $Z$, we assume that the learned hidden representation $S(X)$ contains information about relations between objects.

\subsection{Structured Attention Fusion (SAF) Module}
\label{sec:SIF}
% The goal of the SAF module is to enable the exchange of information between the VI and SI pathways during training. We achieve this by allowing the  SI pathway to attend to the VI pathway. By doing so, structured representation can be further refined in context of learned visual cues from the VI pathway. The opposite is true and we also allow for the visual information to attend to the features from the structured information pathway. 

The objective of the SAF module is to fuse time-series structure of an action captured by the SI pathway and the visual cues from the VI pathway. We formulate the SAF module as a self-attention between the representations from the two pathways. The intuition is that the attention operation adds contextual information learned from video to structural information of an action extracted from objects. We also model the opposite direction of information flow such that the SAF module adds structural information of an action to the visual features.

Let the intermediate values $s_L(Z(X)) \in \mathbb{R}^{T_S^L \times C_S^L}$ and $v_L(X) \in \mathbb{R}^{D_v \times C_V^L}$ from both pathways be inputs to a SAF module. Note that $v_L(X)$ can be a tensor resulting from a flattening operation over the spatial-temporal dimensions such that $D_v = H_V^LW_V^LT_V^L$ or an output of some dimension reduction method (eg. pooling). The SAF operator produces two fused outputs $F_S,F_V$ given $s_L(Z(X)),v_L(X)$. For notation clarity, we abbreviate the terms $s_L(Z(X)), v_L(X)$ as $S, V$ from now on. We define the SAF module as an attention operator between the two pathways. Let us first consider $S$ attending to $V$ denoted as $A_{S \shortrightarrow V}$:
\begin{equation}
\label{eq:sif1}
A_{S \shortrightarrow V}  =  \softmax( \frac{(SW_S)(W^T_VV^T)}{\sqrt{C}})VW_V
\end{equation}
where $W_V \in \mathbb{R}^{C_V^L \times C}$ and $W_S \in \mathbb{R}^{C_S^L \times C}$ are learnable parameters of the SAF module and $C$ is a hyperparameter. Then, the output $F_S$ is computed as:
\begin{equation}
    F_S = \sigma(A_{S \shortrightarrow V})W_{sy} + S
\end{equation}
where $\sigma$ is a normalization operation followed by a non-linearity and $W_{sy} \in \mathbb{R}^{C \times C_S^L}$ is another learnable transformation. By allowing $S$ to attend to $V$ and adding the attended information back to itself, $F_S$ contains information that is fused from both pathways and can be consumed by other operations in the SI pathway. Similarly, we compute $F_V$ by allowing $V$ to attend to $S$ such that:
\begin{equation}
\begin{split}
\label{eq:sif2}
A_{V \shortrightarrow S}  =&  \softmax( \frac{(VW_V)(W^T_SS^T)}{\sqrt{C}})SW_S \\
F_V =& \sigma(A_{V \shortrightarrow S})W_{vy} + V
\end{split}
\end{equation}
where $W_{vy} \in \mathbb{R}^{C \times C_V^L}$. A visual illustration of the SAF module is shown in Figure \ref{fig:saf}. The SAF module can be inserted between the two pathways in different ways to encode different assumptions about the fused representation.
 
%\subsection{Instantiations}

\section{Experiments on Something-Else}
We use the Something-Else \cite{Materzynska_2020_CVPR} task for validating our approach for recognizing compositional actions. 
%Many actions found in the Something-Else dataset can be described by fine-grained changes in geometric relations between the actors in the scene, often including an agent's hand interacting with numerous daily objects. 
The Something-Else task is an extension to the Something-Something-V2 \cite{something} dataset with new object annotations and compositional action recognition splits. The dataset contains a total of 174 action categories where a crowd-sourced worker uploads a video capturing an arbitrary composition of an action category (verb) with an object (noun). As a result, the data set contains a very diverse set of verb-noun compositions involving 12,554 different object descriptions. In Sections \ref{sec:exp_comp} and \ref{sec:exp_fewshot}, we discuss more in detail the compositional and fewshot compositional action recognition tasks. 

We first provide details necessary for reproducing our results in Section \ref{sec:implementation}. We then follow with our main experimental results from the compositional split in Section \ref{sec:exp_comp} and the few-shot split in Section \ref{sec:exp_fewshot}. Finally, we perform thorough ablations on different components of the framework in Section \ref{sec:ablations}.  

\subsection{Implementation Details}
\label{sec:implementation}
Our framework is general and can thus support most recent state of the art approaches for the VI and SI pathways. We detail our choices below.

\noindent \textbf{VI Pathway. } The VI pathways is based on the TSM \cite{tsm} model because of its efficiency and validated performance on the Something-Something-V2 dataset. We use ResNet-50 \cite{resnet} for the backbone and initialize with ImageNet \cite{ILSVRC15} pretrained weights. As standard, we use a base learning rate of 0.01 and a batch size of 64 without a separate warm up schedule. We decrease the learning rate by a factor of 10 at epochs 20 and 40. Training is terminated at 50 epochs. We use stochastic gradient descent with a momentum term with weight 0.9 and a weight-decay of 0.0005.  All settings are adjusted consistently as proposed in \cite{tips} when a change to the batch size was inevitable due to hardware limitations. All our experiments are implemented on the Pytorch framework \cite{pytorch} using 8 Titan Xp GPUs.

\noindent \textbf{SI Pathway. } The SI pathway is a flexible concept and we can implement it using any temporal model including a temporal conv-net (TCN) \cite{restcn,tcn}, a recurrent architecture \cite{gru,lstm} or a recent Transformer encoder model \cite{Vaswani_attention}. We use a TCN because it is efficient and easy to train.
%We adopt a feed forward neural network with repeated 1D convolutions (TCNs) similar to \cite{tcn,restcn} for reasons of efficiency and its natural fit with our framework. Standard RNN-based architectures such as \cite{lstm,gru} as well as recent Transformer encoder models \cite{Vaswani_attention} are reasonable alternatives. 
All training conditions are consistent with the VI pathway. We report ablations on the architectural design choices in Section \ref{sec:ablations}. The details necessary for reproducing our results including SI pathway implementation can be found in the supplementary material.

\noindent \textbf{SI Pathway Input. } We use object detections to form the structured input $Z(X)$ for the SI pathway. Given the $t$-th frame of video $X$ with $N$ objects, we represent a frame as a concatenation of object coordinates such that $Z(X(t))= (x_1^1,y_1^1,x_2^1,y_2^1, \dots, x_1^N,y_2^N,x_2^N,y_2^N,) \in \mathbb{R}^{4N}$ where $(x_1^n,y_1^n,x_2^n,y_2^n)$ are the bounding-box coordinates of the $n$-th object. Then, the video is represented as a temporal concatenation such that $Z(X) = [Z(X(1), \dots, Z(X(T)] \in \mathbb{R}^{4N \times T}$

We construct $Z(X)$ using both ground-truth object locations and predictions obtained from a fine-tuned object detector \cite{faster_rcnn}. For fair comparison, we use the same tracking results provided by the authors of \cite{Materzynska_2020_CVPR} to establish temporal correspondence between object predictions.

\subsection{Compositional Action Recognition}
\label{sec:exp_comp}
In the original action recognition split of Something-Something-V2, the same verb-noun composition may appear in both training and validation sets. We first describe the compositional action recognition split \cite{Materzynska_2020_CVPR} that prevents this.

\noindent \textbf{Problem formulation.}
Let there be two disjoint sets of nouns (objects), $\{ \mathcal{A}, \mathcal{B} \}$, and two disjoint sets of verbs (actions) $\{1,2\}$. The goal of the compositional action recognition task is to recognize novel verb-noun compositions at test time. The model can observe instances from the set $\{1\mathcal{A}+2\mathcal{B}\}$ during training but will be tested using instances from $\{1\mathcal{B}+2\mathcal{A}\}$. In this setting, there are 174 action categories with 54,919 training and 57,876 validation instances.

%% GT

\begin{table}[t]
\centering
\begin{tabular}{l|cc|cc}
                \multicolumn{1}{c|}{Model} & \multicolumn{2}{c|}{Input} & \multicolumn{2}{c}{Evaluation} \\
 & RGB         & Objects         & top-1      & top-5\\ \hline
I3D \cite{Carreira2017QuoVA} & o  & & 46.8 & 72.2  \\
TSM \cite{tsm} & o & & 52.3 & 78.0 \\ \hline
STIN \cite{Materzynska_2020_CVPR}& & o & 51.4&79.3\\
STIN  & o & o & 54.6 &79.4 \\
STIN ensemble    & o & o & 58.1 &83.2\\ \hline
Ours (SI Only) &  & o & 53.8 & 79.8\\
Ours & o & o & \textbf{60.5} & \textbf{84.3} \\
Ours ensemble & o & o & \textbf{67.3} &\textbf{ 89.3}          
\end{tabular}
\caption{Results on compositional action recognition using \textbf{ground-truth} objects.}
\label{tab:comp_GT}
\end{table}

%% DETECTIONS
\begin{table}[t]
\centering
\begin{tabular}{l|cc|cc}
                \multicolumn{1}{c|}{Model} & \multicolumn{2}{c|}{Input} & \multicolumn{2}{c}{Evaluation} \\
 & RGB         & Objects         & top-1      & top-5\\ \hline
I3D \cite{Carreira2017QuoVA} & o  & & 46.8 & 72.2  \\
TSM \cite{tsm} & o & & 52.3 & 78.0 \\ \hline
STIN  \cite{Materzynska_2020_CVPR}  & o & o & 48.2 & 72.6 \\ 
STRG \cite{Wang_videogcnECCV2018} & o & o & 52.3 & 78.3 \\
Ours  & o & o &  \textbf{60.7} & \textbf{84.2}  \\ \hline
STIN ensemble \cite{Materzynska_2020_CVPR}  & o & o & 51.5 & 77.1 \\ 
STRG ensemble\cite{Wang_videogcnECCV2018} & o & o & 56.2 & 81.3 \\
Ours ensemble & o & o &  \textbf{65.3} & \textbf{87.9}
\end{tabular}
\caption{Results on compositional action recognition using \textbf{detected} objects. }
\label{tab:comp_DET}
\end{table}

\noindent \textbf{Baselines. } For RGB-only baselines, we use the popular I3D \cite{Carreira2017QuoVA} model and the state-of-the-art (on Something-Something-V2) TSM  \cite{tsm}. For methods that use both RGB and object bounding boxes, we compare to a very recent state-of-the-art Spatial-Temporal Interaction Network (STIN) \cite{Materzynska_2020_CVPR} that models actions as sparse and semantically-rich object graphs. We compare to the Space-Time Region Graph (STRG) model as well when comparable results are available.

\noindent \textbf{Results. } In Table \ref{tab:comp_GT}, we first report our experiments using ground truth object detections to isolate the effect of object detector performance from the effect of our approach. When accurate object locations are known, we show that simple baselines using only object-level information outperform video models.
%We show that without any notion of structural information, highly parameterized baselines such as the I3D performs worse than much simpler models that only use object bounding boxes without any visual input. The finding is consistent even when comparing against much stronger RGB only baselines such as the TSM.
This shows that popular RGB-only baselines do not generalize over object appearance and/or fail to capture structural information in the scene, and thus does not perform well when presented with novel verb-noun compositions at test time. 

When using videos in conjunction with object detections, our approach of fusing information from the SI and VI pathways using the SAF module improves the current state-of-the-art results by a considerable margin. As seen in Table \ref{tab:comp_GT}, the benefit of using RGB video in addition to object detections is $3.2\%$ for STIN whereas SAFCAR improves $6.7\%$ --  more than double the positive gain. Although SAFCAR and STIN perform similarly on the top-5 evaluation ($79.8\%$ vs. $79.3\%$) when only using object detections, there is a significant performance gain of $4.5\%$ when visual information is fused using our approach. This compares to a minimal improvement of $0.1\%$ for STIN. 
%This is an interesting finding as it indicates that our approach uses visual information more effectively to resolve reasonable `near-misses'. 
We show that a simple ensemble of predictions (combining RGB-only, an object-only and a RGB+object model, all independently trained) leads to additional performance gains where we establish a strong new state-of-the-art performance of $67.3\%$  top-1 (an increase of $9.2$ points) and $89.3\%$ top-5 (an increase of $6.1$ points) accuracies.  

Table \ref{tab:comp_DET} shows action recognition performance on the same compositional split when using detected objects instead of ground truth annotations. While all methods perform worse with object predictions, the overall findings are consistent with the experiments reported in Table \ref{tab:comp_GT} -- namely, our approach outperforms other methods whether ensemble-ed with baselines or not.

\subsection{Few-shot Compositional Action Recognition}
\label{sec:exp_fewshot}

\begin{table}[]
\centering
\begin{tabular}{l|cc|cc}
                \multicolumn{1}{c|}{Model} & \multicolumn{2}{c|}{Input} & \multicolumn{2}{c}{Evaluation} \\
 & RGB  & Objects  & 5-shot & 10-shot     \\ \hline
I3D \cite{Carreira2017QuoVA} & o & & 21.8 & 26.7 \\
TSM \cite{tsm} & o & & 22.5 & 27.3 \\ \hline
STIN \cite{Materzynska_2020_CVPR}& & o & \textbf{27.7} & \textbf{33.5}  \\
STIN & o & o & 28.1 & 33.6 \\
STIN ensemble & o & o & 34.0 & 40.6 \\ \hline
Ours (SI only) & & o & 27.0& 32.8 \\
Ours  & o & o & \textbf{31.5} & \textbf{36.8}\\
Ours ensemble & o  & o & \textbf{39.0} &  \textbf{45.4}
%\\ \hline
% Ours-L &  & o & \textcolor{blue}{29.3 (29.7)} & \textcolor{blue}{36.0 (36.9)}  \\
% Ours-L  & o & o & \textcolor{blue}{31.1 (R)} & \textcolor{blue}{36.1} \\
% Ours-L ensemble & o & o & \textcolor{blue}{38.8}  &  \textcolor{red}{TODO}           

\end{tabular}
\caption{Results on few-shot compositional action recognition using \textbf{ground-truth} objects}
\label{tab:fewshot_GT}

%% DETECTION

\begin{tabular}{l|cc|cc}
                \multicolumn{1}{c|}{Model} & \multicolumn{2}{c|}{Input} & \multicolumn{2}{c}{Evaluation} \\
 & RGB  & Objects  & 5-shot & 10-shot     \\ \hline
I3D \cite{Carreira2017QuoVA} & o & & 21.8 & 26.7 \\
TSM \cite{tsm} & o & & 22.5 & 27.3 \\ \hline

STIN \cite{Materzynska_2020_CVPR}& & o & 17.7 & 20.7  \\
Ours (SI only)& & o & 15.8 & 19.8 \\\hline
STIN & o & o & 23.7 & 27.0 \\
STRG \cite{Wang_videogcnECCV2018} & o & o & 24.8 & 29.9
\\
Ours  & o & o & \textbf{28.5} & \textbf{32.0} \\ \hline
STIN ensemble & o & o & 27.3 & 32.6  \\
STRG ensemble & o & o & 29.1 & 34.6 \\ 
Ours ensemble  & o & o & \textbf{37.0} & \textbf{43.4} \\\hline

\end{tabular}
\caption{Results on few-shot compositional action recognition using \textbf{detected} objects} 
\label{tab:fewshot_DET}
\end{table}

\noindent \textbf{Problem formulation.}
In the compositional split, a common set of all 174 action categories appear during training and testing of the model. In the \textit{few-shot}  compositional split, the original 174 action categories are divided into a \textit{base} set with 88 categories and a \textit{novel} set with 86 action categories. The \textit{base} set contains 112,397 labeled training instances where the model is `pretrained'. Then, we only use few labeled examples (fewshot-train) from each novel category to `finetune' the model to recognize rest of the videos from the novel action categories (fewshot-validation). For example, when performing 5-shot transfer, only $5 \times 86 = 430$ labeled instances are available during the finetuning stage. As with the compositional split, the objects that appear in fewshot-train are independent from the objects that appear in the fewshot-validation. There are 49,822 and 43,957 instances in the 5-shot and 10-shot validation sets respectively. We pretrain a model using the base categories and freeze the trained parameters when finetuning with the few-shot examples.

\noindent \textbf{Results}
For a model to perform well under this setting, it must learn re-usable spatial-temporal structures during pretraining with available aciton categories such that it can quickly adapt to unseen actions (verbs) during finetuning. In Table \ref{tab:fewshot_GT}, we compare our approach to other state of the art methods for this task that use ground-truth object bounding boxes as part of the input. We observe that both RGB only baselines (I3D,TSM) perform poorly which further underscores that they struggle to generalize over appearance of objects when given novel verb-noun compositions. Compared to the compositional split results in Table \ref{tab:comp_GT} where RGB-only baselines performed on par with object-only models, RGB-only baselines actually perform worse than models that only use object level information in the few-shot setting. 

The few-shot experiments clearly highlight the benefit of our approach in making a better use of the visual information when learning with objects. Compared to other state of the art baselines, the performance boost induced by the visual content is large. The object-only baselines of both our approach and that of the STIN model perform similarly: $27.7\%$ vs. $27.0\%$ for the 5-shot and $33.5\%$ vs. $32.8\%$ for the 10-shot splits. When the models are allowed to observe RGB videos, the STIN model gains $0.4$ for the 5-shot setting and $0.1$ for the 10-shot setting. In contrast, we gain $4.5$ and $4.0$ points respectively. This is a strong evidence that our approach fuses visual information more effectively.

We compare results when using object detections instead of ground truth object annotations in Table \ref{tab:fewshot_DET}. Our SI only model is actually weaker than other state of the art baselines. However, when using video data in addition to object detection, we observe a significant improvements in performance of $12.7\%.$ In comparison, the STIN model gains only $6\%$.

\subsection{Ablations}
\label{sec:ablations}
We perform ablation studies on the few-shot compositional split of the Something-Else task.

\begin{figure}[t]
\centering
\includegraphics[width=1.0\linewidth]{ 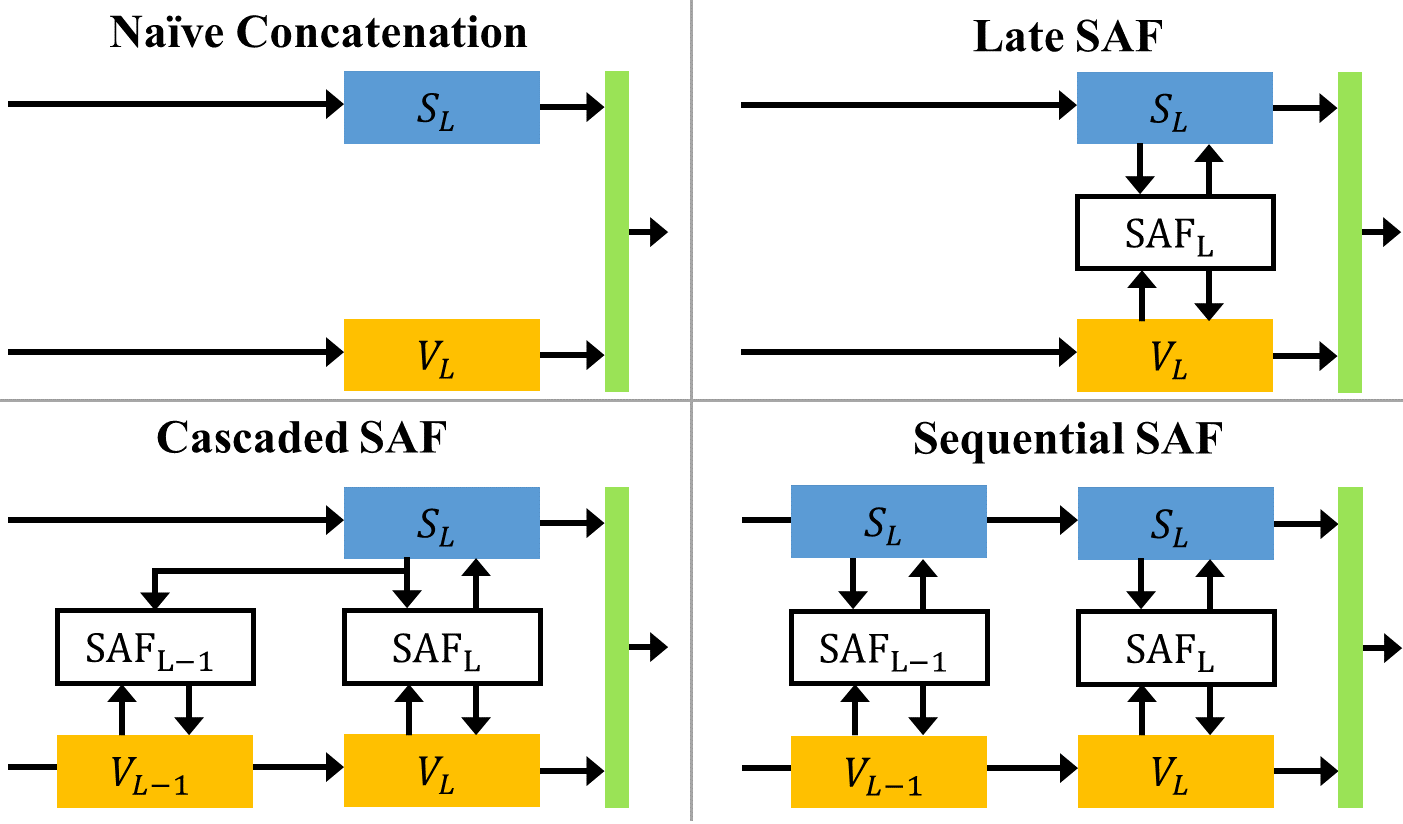}
  \caption{A visual comparison of different ways the SAF module can be inserted between the SI and VI pathways.}
\label{fig:SAF_ablate}
\end{figure}

\noindent \textbf{SAF architectures: } Table \ref{tab:saf} shows various ways of using the SAF module. A simple concatenation of the outputs from both pathways improves the performance of the VI-only baseline. However, naive concatenation actually performs worse than the model using only object detections (SI-only baseline). We ablate our approach with three different ways of using the SAF module as illustrated in Figure \ref{fig:SAF_ablate}. The Late-SAF version inserts a SAF module between the final feature extraction layers of both pathways to fuse high-level information.
%In the Late-SAF instantiation, the SAF module is fusing high-level information between the pathways.
The Cascaded-SAF version allows the high-level structural information to affect the lower level visual feature learning process. Finally, Sequential-SAF fuses intermediate layers sequentially. Table \ref{tab:saf} shows that all variations that use the SAF module perform better than when the information from both pathways are naively merged. We also observe that fusing high-level information (Late-SAF) is more effective than other variants. 

\noindent \textbf{SI pathway architectures: } Table \ref{tab:si_ablation} shows different SI pathway implementations. All versions are based on 1D convolution layers with filter length 3 and a stride of 1. Models v0 and v1 share identical temporal convolutional layers but v1 contains extra multi-headed attention transformer encoder \cite{Vaswani_attention} layers before the prediction head. Model v2 is a purely convolutional architecture with residual connections \cite{resnet,restcn}. The architecture of v3 is the same as v2 but with more convolutional filters per layer to contain almost double the number of trainable parameters. We see that adding parameters improves performance when only using objects as input. However, in the next ablation study, we show that increasing the complexity of the SI pathway does not necessarily lead to improved few-shot performance when combined with the VI pathway.

\begin{table}[t]

    % SI-pathway ablation
    % \begin{subtable}[h]{0.45\textwidth}
    %     \centering
    %     \begin{tabular}{c|cccc}
    %     SI Pathway & C-top1 & FS-5 & FS-10 & \# Params \\ \hline
    %     v1 & 54.4 & 27.0 & 32.8    &    5.79M  \\
    %     v2 & 53.8 & 29.3 & 36.0    &    5.27M  \\
    %     v3 & 55.4 & 29.7  & 36.9 & 9.23M     
    %     \end{tabular}
    %   \caption{SI Pathway instantiations. Results when using only objects.}
    %   \label{tab:si_ablation}
    % \end{subtable}
    % \hfill
    %     \vspace{0.1cm}
        
     % SI-pathway ablation
     % SAF ablation
     %%%%%%%%%%%%%%%%5
    \begin{subtable}[h]{0.45\textwidth}
        \centering
        \begin{tabular}{c|cc}
\multicolumn{1}{c|}{} & 5-shot & 10-shot \\ \hline
Baseline (VI only) & 22.5 & 27.3 \\
Baseline (SI only) & 27.0 & 32.8 \\
Concat(SI,VI) (No-SAF) & 25.6 & 32.1 \\\hline
Late-SAF    & \textbf{31.5} & \textbf{36.8} \\
Cascaded-SAF & 29.9 & 34.8\\
Sequential-SAF & 30.4& 35.6
\end{tabular}
        \caption{Different ways to use the SAF module.}
        \label{tab:saf}
     \end{subtable}
         \hfill
         \vspace{0.50cm}
         
         %%%%%%%%%%%%%%%%%
         
    \begin{subtable}[h]{0.45\textwidth}
        \centering
        \begin{tabular}{c|ccc}
        SI Pathway Model &  FS-5 & FS-10 & \# Params \\ \hline
        v0 & 26.4 & 31.2    &    5.19M  \\
        v1 & 27.0 & 32.8    &    5.79M  \\
        v2 & 29.3 & 36.0    &    5.27M  \\
        v3 &  29.7  & 36.9 & 9.23M     
        \end{tabular}
       \caption{Different SI pathway instantiations using only ground truth detections.}
       \label{tab:si_ablation}
    \end{subtable}
    \hfill
        \vspace{0.5cm}

  % SI+VI pathway ablation
    \begin{subtable}[h]{0.45\textwidth}
        \centering
        \begin{tabular}{c|cc}
            Combination & FS-5 &FS-10  \\ \hline
            v0 + Late-SAF  & 28.2 & 33.7 \\
            v1 + Late-SAF  & \textbf{31.5} & \textbf{36.8} \\
            v2 + Late-SAF  & 31.1 & 36.1 \\
            v3 + Late-SAF  & 29.4 & 34.3 \\
            \end{tabular}
      \caption{Different SI pathway instantiations combined with the the same VI pathway.}
      \label{tab:si_vi_combination_ablation}
    \end{subtable}
    \hfill
    \vspace{0.1cm}
    
     \caption{Ablations using the few-shot compositional split.}
     \label{tab:ablations}
\end{table}
\noindent \textbf{SI+VI pathway combinations: }
Table \ref{tab:si_vi_combination_ablation} shows the performance of different SI pathway implementations when combined with a constant VI implementation. We find that adding the transformer encoder layers (v0 vs. v1) in the SI pathway improves performance when fused with the VI pathway. From Table \ref{tab:si_ablation}, we saw that v2 outperforms v1 when only object input is used. However, we find that when combined with the VI pathway, both models perform similarly. Lastly, we begin to see signs of overfitting with v3.

\section{Experiments on Charades-Fewshot}
To demonstrate that our framework for recognizing compositional actions is flexible, we evaluate our method on the Charades \cite{sigurdsson2016hollywood} dataset which contains 9,848 videos and 157 action categories with an average length of 30 seconds which may contain one or more action instances. Many action categories found in the Charades dataset are defined as a verb+noun composition (ie. Opening a book, Opening a door, Throwing a book). The task is to predict all the actions contained in the video without having to predict the start and end frames of the instances.
Despite the different set of activities and objects, we show that our approach generalizes effectively across domains and show convincing few-shot compositional action recognition capabilities on the Charades-Fewshot task \cite{Ji_2020_CVPR}.

\subsection{Implementation Details}
Our goal with the Charades dataset is to demonstrate the flexibility of our framework when applied to a different problem domain. Hence, we do not perform any architecture search and simply use the same SI-pathway architecture (v1) used in Section \ref{sec:implementation}. For the VI pathway, we use the SlowFast  \cite{feichtenhofer2018slowfast} model given its competitive performance on the Charades dataset. We finetune the model with ResNet-50 backbone pretrained on Kinetics-400 \cite{kinetics}. Similar to the Something-Else experiments, the input to the SI pathway is a time-series of ground-truth object locations. There are 36 annotated objects such that $Z(X) \in \mathbb{R}^{36*4 \times T}$. Detailed parameters necessary for reproducing our results are included in the supplementary material.

\subsection{Few-shot Action Recognition}

\begin{table}[]

%% DETECTION
\centering
\begin{tabular}{l|ccc}
 & 1-shot& 5-shot & 10shot \\ \hline
LFB \cite{lfb2019} &28.3&36.3&39.6 \\ 
SGFB \cite{Ji_2020_CVPR} & 28.8&37.9&42.7\\
SGFB-oracle \cite{Ji_2020_CVPR} &\textbf{30.4} & 40.2 & 50.5\\ \hline
Ours, VI-only (SlowFast \cite{feichtenhofer2018slowfast}) & 18.9 & 31.1 & 36.4 \\
Ours & 21.8 & \textbf{41.8} & \textbf{50.9}\\
\end{tabular}
\caption{Results on the  Charades-Fewshot task.} 
\label{tab:charades}
\end{table}

\noindent \textbf{Problem Formulation}
The Charades dataset contains 37 annotated objects. Following the Charades-Fewshot setup \cite{Ji_2020_CVPR}, six `novel' objects (pillow, broom, television, bed, vacuum, refrigerator) are excluded during training. There are 20 action categories that involve an interaction with an object in the novel-object set. Therefore, there are 137 base action categories where we pretrain our model and 20 novel actions that are held out. From the 20 novel categories, a fixed set of few (k=1,5,10) examples are sampled per category to form the fewshot-train set where our model is finetuned. The remaining instances are used for few-shot validation. 

\noindent \textbf{Baselines.} We compare our approach to recent state-of-the-art methods that reported results on the Charades-Fewshot task, including the Long-term Feature Banks (LFB) \cite{lfb2019} and Scene-Graph Feature Banks (SGFB) \cite{Ji_2020_CVPR}. LFB extracts ROI-aligned features using object detection as the feature bank representation which gets merged with a video feature for action prediction. SGFB uses additional manual annotations collected using the Action Genome \cite{Ji_2020_CVPR} framework which contains ground truth relations between objects and attribute level information. The annotations include relational attributes such as `in-front-of', `is-looked-at', 'being-touched' and more.
The annotated spatio-temporal scene graphs are provided to the SGFB model as an additional supervision during training. The SGFB-oracle model assumes correct spatial-temporal scene graphs are available during test time.

\noindent \textbf{Results.} We note that all state of the art models for this task use additional information such as objects (LFB) or annotated scene-graphs (SGFB). As expected, our baseline approach using only RGB videos trails other methods in all few-shot settings. Consistent with our findings in the Something-Else task in the previous sections, we show strong results when RGB information is combined with the SI-pathway using the SAF module. We show large improvements over our baseline: we gain 2.9, 10.7 and 14.5 points in 1-shot, 5-shot and 10-shot experiments respectively. We observe that our approach does not sufficiently model the high-level structure of an action using only one example per category when compared to other state of the art baselines. In the extreme 1-shot case, the SGFB-oracle model which assumes knowledge of manually annotated object-level attributes outperforms other methods. When enough examples are available (K $\geq$ 5), we empirically show our approach can even outperform the SGFB-oracle model.  This finding suggests that the SI-pathway of our approach implicitly learns similar high-level concepts that are defined and annotated in the scene-graphs representation of the SGFB-oracle model.
%We empirically show that when using accurate object detections, our approach even outperforms the SGFB-oracle model when enough labeled examples are available (K $\geq$5). High-level structure of an action is given to the SGFB-oracle model   This finding suggests that the SI-pathway of our approach can implicitly learn high-level structure of an action more effectively even without having to manually specify relational attributes.

\section{Conclusion}
We have presented a general framework that uses Structured and Visual pathways with the SAF module. Using object detections as input to the SI pathway, our approach modeled structures of actions better when attending to visual features in the VI pathway via the SAF mechanism. We validate our approach on multiple tasks including the Something-Else (compositional, few-shot compositional splits) and the Charades-Fewshot tasks where we significantly improved over the current state of the art.

\section*{Acknowledgement}
We thank Jin Bai for his consultations on object detectors and tracking. We thank Jonathan Jones for useful discussions.

{\small
\bibliographystyle{ieee_fullname}
\bibliography{cvpr}
}

\end{document}